\pdfoutput=1

\documentclass[11pt]{article}

\usepackage[preprint]{acl}

\usepackage{times}
\usepackage{latexsym}

\usepackage[T1]{fontenc}

\usepackage[utf8]{inputenc}
\usepackage{amsmath}
\usepackage{amssymb}
\usepackage{amsfonts}
\usepackage{bm}
\usepackage{enumitem}
\usepackage{algorithm}
\usepackage{algorithmicx}
\usepackage{algpseudocode}
\usepackage{wrapfig}

\usepackage{bbding}
\usepackage{tabularx}

\usepackage{amsmath}
\usepackage{amssymb}
\usepackage{booktabs}      
\usepackage{multirow}      
\usepackage{xcolor}        
\usepackage{graphicx}      
\usepackage{caption}       
\usepackage{subcaption}    

\usepackage{microtype}

\usepackage{inconsolata}

\usepackage[T1]{fontenc}
\usepackage{colortbl}
\usepackage{booktabs,multirow}
\usepackage{makecell}
\usepackage{tabulary}
\usepackage{fontawesome5}
\usepackage{pifont}
\usepackage{tikz}

\newlength\savewidth\newcommand\shline{\noalign{\global\savewidth\arrayrulewidth
  \global\arrayrulewidth 1pt}\hline\noalign{\global\arrayrulewidth\savewidth}}

\title{Shifting AI Efficiency From \\ Model-Centric to Data-Centric Compression}

\author{%
  Xuyang Liu$^{1,2}$\thanks{Equal contribution: liuxuyang@stu.scu.edu.cn $^\text{\Envelope}$Corresponding author: zhanglinfeng@sjtu.edu.cn} \quad Zichen Wen$^{1,3,4*}$ \quad Shaobo Wang$^{1*}$ \quad Junjie Chen$^1$ \quad Zhishan Tao$^1$ \\ \textbf{Yubo Wang$^1$} \quad \textbf{Tailai Chen$^1$} \quad \textbf{Xiangqi Jin$^{1,3}$} \quad \textbf{Chang Zou$^{1,3}$} \quad \textbf{Yiyu Wang$^1$} \quad \textbf{Chenfei Liao$^6$} \\ \textbf{Xu Zheng$^6$} \quad \textbf{Honggang Chen$^2$} \quad \textbf{Weijia Li$^{4,5}$} \quad \textbf{Xuming Hu$^6$} \quad \textbf{Conghui He$^4$} \quad \textbf{Linfeng Zhang$^{1\text{\Envelope}}$} \\
    $^1$EPIC Lab, Shanghai Jiao Tong University \quad
    $^2$Sichuan University \\
    $^3$University of Electronic Science \& Technology of China \quad $^4$Shanghai AI Laboratory  \\ $^5$Sun Yat-sen University \quad $^6$Hong Kong University of Science and Technology (Guangzhou) \vspace{3pt}\\
  \textbf{Project: \href{https://github.com/xuyang-liu16/Awesome-Token-level-Model-Compression}{\texttt{\textcolor{cyan}{Awesome-Token-level-Model-Compression}}}}
}

\begin{document}

\maketitle

\begin{abstract}

The advancement of large language models (LLMs) and multi-modal LLMs (MLLMs) has historically relied on scaling model parameters.
However, as hardware limits constrain further model growth, the primary computational bottleneck has shifted to the quadratic cost of self-attention over increasingly long sequences by ultra-long text contexts, high-resolution images, and extended videos.
In this position paper, \textbf{we argue that the focus of research for efficient artificial intelligence (AI) is shifting from model-centric compression to data-centric compression}. 
We position data-centric compression as the emerging paradigm, which improves AI efficiency by directly compressing the volume of data processed during model training or inference.
To formalize this shift, we establish a unified framework for existing efficiency strategies and demonstrate why it constitutes a crucial paradigm change for long-context AI.
We then systematically review the landscape of data-centric compression methods, analyzing their benefits across diverse scenarios. Finally, we outline key challenges and promising future research directions. Our work aims to provide a novel perspective on AI efficiency, synthesize existing efforts, and catalyze innovation to address the challenges posed by ever-increasing context lengths.

\end{abstract}

\section{Introduction}
\label{sec:introduction}

\begin{figure*}
    \centering
    \includegraphics[width=1.0\textwidth]{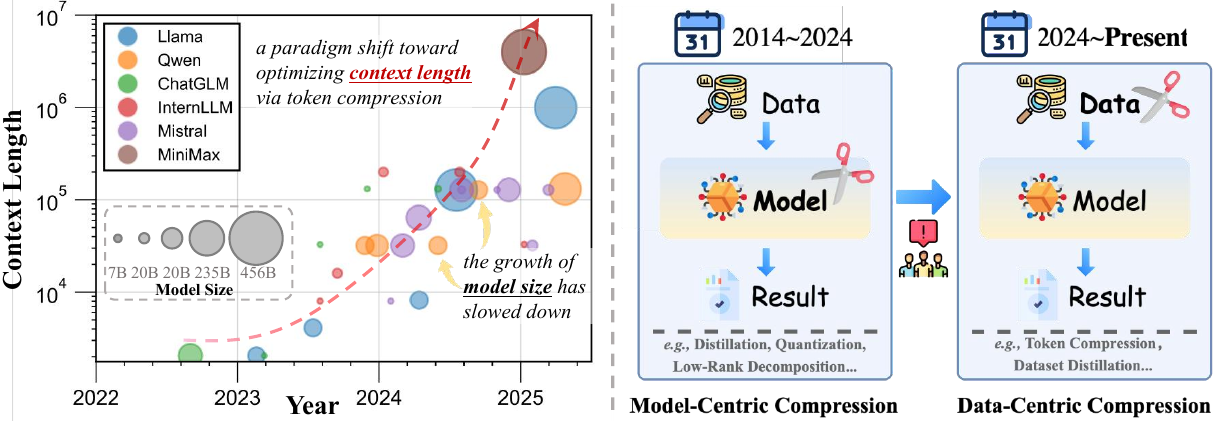}
    \vspace{-7mm}
    \caption{\textbf{The evolution of AI efficiency: from model-centric to data-centric compression.} From 2022 to 2024, AI model performance gains were primarily driven by scaling \emph{model size}, directing efficiency research toward \textbf{\emph{model-centric compression}}. By 2024, with model sizes approaching 1T parameters, their growth has \textcolor[RGB]{248,170,52}{\textbf{slowed down}}. Consequently, the focus has shifted to \textcolor[RGB]{223,52,39}{\textbf{expanding}} \emph{context length} to enhance model capabilities, necessitating a transition to \textbf{\emph{data-centric compression}} that reduces context length for efficiency.}
    \label{fig:evolution}
    \vspace{-5mm}
\end{figure*}

The explosive growth of large language models (LLMs)~\cite{OpenAI:GPT-4,Touvron:LLaMA,grattafiori2024llama3,dong2024internlm,yang2025Qwen3,guo2025deepseek-r1} and their multi-modal extensions (MLLMs)~\cite{Liu:LLaVA-1.5,Chen:InternVL,zhu2025internvl3,wang2024qwen2-vl,bai2025qwen2.5-vl} over the past few years has driven remarkable gains in AI capabilities. This progress has been largely achieved by increasing \emph{model scale}, with larger models consistently showing superior performance in reasoning, knowledge acquisition, and task generalization. The evolution from early models like BERT (117M)~\cite{devlin2018bert} to today's state-of-the-art LLMs such as DeepSeek-R1~\cite{guo2025deepseek-r1} and Qwen-3~\cite{yang2025Qwen3} (100B+) illustrates how scale has delivered substantial performance improvements. Nevertheless, this pursuit of performance through larger models incurs ever-increasing computational costs. As a result, by early 2024, the dominant source of computational overhead was primarily attributed to the \emph{linear growth in parameter count and associated memory requirements}.

In response to this scaling trend, the research community has developed numerous \textbf{\emph{model-centric compression}} techniques, including model quantization~\cite{yang2019quantization,rokh2023comprehensive}, network pruning~\cite{Deep_Compression,cheng2024survey}, knowledge distillation~\cite{Hinton:KD,gou2021KD_Survey}, and low-rank decomposition~\cite{yu2017low-rank,idelbayev2020low-rank}. These methods reduce computational overhead by decreasing model size and were a natural response to the 2022–2024 era, when scaling model size was the primary driver of performance gains.

As model sizes approach hardware limits, the pace of parameter growth is flattening. Meanwhile, a new computational challenge has emerged: the exponential growth in \emph{context sequence lengths}. Figure~\ref{fig:evolution} (left) shows that from 2022 to 2024, model size primarily drove computational costs, reaching around 1T parameters before stagnating. Since then, the dominant factor has shifted dramatically to the staggering number of processed tokens, which continues to grow exponentially. This trend spans multiple domains: language models now handle context lengths orders of magnitude longer than before~\cite{llama4,yang2025Qwen3}, especially with long chain-of-thought reasoning~\cite{guo2025deepseek-r1} and multi-agent systems~\cite{han2024llm-multi-agent}; vision models process increasingly high-resolution images~\cite{zhu2025internvl3,bai2025qwen2.5-vl} and longer videos~\cite{qin2025video-xl-2}; and generative models create higher-resolution images~\cite{flux2024} and hour-long videos~\cite{openai2024sora}, all requiring more tokens and causing substantial computational overhead. Consequently, by late 2024, the primary bottleneck has clearly shifted to the \emph{quadratic cost of the attention mechanism over these extremely long context sequences}.

This unprecedented growth in sequence lengths has shifted the computational bottleneck from model size to the quadratic cost of attention over long context sequences. Based on this observation, as illustrated in Figure~\ref{fig:evolution} (right), we propose a critical position: \textbf{the AI community should shift its efficiency optimization paradigm from model-centric to data-centric compression}. We advocate for \textbf{\emph{data-centric compression}} that directly reduces the volume of data processed during model training or inference~\cite{jiang2023llmlingua,Bolya2023:ToMe,Bolya2023:ToMeSD,Chen2024:FastV,rho1}. These methods address computational overhead by removing low-information content during processing, typically without modifying model architectures or requiring retraining. Our analysis in Section~\ref{subsec:advantages} shows that they offer compelling advantages in \textbf{universality, efficiency, and compatibility}, positioning data-centric compression as a promising solution for efficient next-generation LLMs and MLLMs.

Building upon these analyses, we make four key contributions in this position paper:

\begin{itemize}
[leftmargin=10pt, topsep=0pt, itemsep=1pt, partopsep=1pt, parsep=1pt]
    \item \textbf{Evolution of AI Efficiency:} We analyze recent developments in long-context AI across various domains, revealing a critical transition from parameter-centric to context-centric computational bottlenecks that necessitate a paradigm shift in efficiency optimization.

    \item \textbf{Unified Formulation of Model Efficiency:} We establish a comprehensive mathematical formulation that unifies architectural design, model-centric compression, and data-centric compression within a single expression.
    
    \item \textbf{Systematic Review of Data-centric Compression:} We present a thorough investigation of data-centric compression methods, constructing a unified framework to categorize diverse approaches while analyzing their benefits across different scenarios and tasks.
    
    \item \textbf{Challenges and Future Directions:} We provide an in-depth analysis of current challenges in data-centric compression research and propose promising future directions, aiming to catalyze research efforts toward more efficient and effective compression methods.

\end{itemize}

\section{Background}
\label{sec:background}
\subsection{Token Overhead across Various Domains}
\label{sec:transformer-based model}


The field of AI has witnessed remarkable advancements across multiple domains, including natural language processing, computer vision, and content generation. These developments have been largely driven by the introduction of the Transformer architecture~\cite{vaswani2017:Transformer}, which has spawned a wide variety of models. As these domains evolve, we observe a significant increase in token sequence lengths across \textbf{\emph{three main areas}}:

\noindent \textbf{(I) Longer Context Length in Language Models:} Large language models (LLMs)~\cite{OpenAI:GPT-4,Touvron:LLaMA,grattafiori2024llama3,liu2024deepseek-v3,Bai:Qwen,yang2025Qwen3} have demonstrated remarkable capabilities in natural language understanding and generation. The context length LLMs can handle has expanded dramatically from 2,048 tokens in early models like Llama 1~\cite{Touvron:LLaMA} to 10M tokens in recent iterations like Llama 4 Scout~\cite{llama4}. This expansion has led to the emergence of large reasoning models~\cite{guo2025deepseek-r1,yang2025Qwen3}, which focus on complex multi-step problem solving through techniques like long chain-of-thought reasoning~\cite{liu2024deepseek} and multi-agent collaboration~\cite{han2024llm-multi-agent}.

\noindent \textbf{(II) Higher Resolution and Longer Video Understanding:} Building on the success of LLMs, multi-modal large language models (MLLMs)~\cite{Liu:LLaVA-1.5,li2024llava-ov,Bai:Qwen-VL,bai2025qwen2.5-vl,Chen:InternVL,zhu2025internvl3,guo2025seed-vl} extend these capabilities by integrating vision and text processing~\cite{wu2023multimodal}. Visual inputs processed by MLLMs have evolved from basic $224 \times 224$ resolution images in early models like LLaVA~\cite{liu2023llava} to 4K ultra-high-resolution images in InternVL3~\cite{zhu2025internvl3} and 10K-frame videos in Video-XL-2~\cite{qin2025video-xl-2}, enabling strong performance on image~\cite{bai2025qwen2.5-vl}, video~\cite{yang2025Keye-VL-1.5}, and multi-modal reasoning tasks~\cite{shen2025vlmr1}.

\noindent \textbf{(III) More Complex Content in Generation Tasks:} AI content generation has advanced significantly with the application of Transformers to generative domains~\cite{peebles2023DiT,openai2024sora,li2024hunyuandit}. Early diffusion models like Stable Diffusion~\cite{rombach2022SD} generated only $512 \times 512$ resolution images. With Transformers now successfully applied to generation~\cite{peebles2023DiT,gao2023maskdit,openai2024sora,li2024hunyuandit}, DiT-based models produce high-quality 4K images in PixArt-$\Sigma$~\cite{chen2024pixart-sigma} and hour-long videos in Sora~\cite{openai2024sora}. These models capture complex spatiotemporal dependencies, enabling high-fidelity content generation~\cite{flux2024,yang2024cogvideox,wan2025wan,kang2025legion}.

While these advancements across domains have demonstrated outstanding performance, they now face significant efficiency challenges due to the \emph{quadratic cost of attention mechanisms over extremely long token sequences}. The growing trend toward longer contexts, whether in complex reasoning chains for language tasks, high-resolution images and longer videos for understanding, or high-fidelity content for generation, necessitates prioritizing research on model efficiency, particularly in mitigating the computational overhead of increasing context lengths. Detailed statistical analysis of this trend is provided in Appendix~\ref{sec:llm_trends}.

\subsection{AI Efficiency from Different Perspectives}
\label{subsec:compression-definition}

Improving model efficiency has been a key goal in deep learning research. Given input data $\mathbf{X}$ and network parameters $\mathbf{W}$, a neural network $\mathbf{F}$ produces output $\mathbf{Y}$ through the transformation:
\begin{equation}
\underbrace{\mathbf{Y}}_{\text{output}} = \underbrace{\mathbf{F}}_{\text{network}}(\underbrace{\mathbf{W}}_{\text{weights}}, \underbrace{\mathbf{X}}_{\text{input}})
\end{equation}
Model efficiency can be optimized from three perspectives: \textbf{(I)} Efficient Computation Architecture designs efficient neural architectures $\mathbf{F}$~\cite{shen2021-liner-attention,peng2023rwkv,gu2023mamba}, \textbf{(II)} Model-centric Compression reduces model weights $\mathbf{W}$~\cite{Hinton:KD,yang2019quantization,li2017pruning,yu2017low-rank}, and \textbf{(III)} Data-centric Compression compresses token sequences from input data $\mathbf{X}$~\cite{Rao2021:DynamicViT,jiang2023llmlingua,Chen2024:FastV,zou2025:ToCa}.

\noindent \textbf{(I) Efficient Computation Architecture ($\mathbf{F}$):} Since computational efficiency is determined by architectural design, optimizing $\mathbf{F}$ is fundamental. Unlike Transformers with \textbf{\emph{quadratic}} attention complexity $\mathcal{O}(n^2)$~\cite{vaswani2017:Transformer}, recent methods achieve \textbf{\emph{linear or sub-quadratic}} scaling: \textbf{(i) linear attention} reformulates attention for $\mathcal{O}(n)$ complexity~\cite{katharopoulos2020:transformers-rnn,shen2021-liner-attention}; \textbf{(ii) RWKV} combines RNN-like $\mathcal{O}(n)$ scaling with transformer parallelism~\cite{peng2023rwkv,duan2024visionrwkv}; \textbf{(iii) State Space Models} like Mamba use structured state spaces for $\mathcal{O}(n)$ complexity~\cite{gu2023mamba,Zhu2024:Vim}. These require retraining, motivating alternative approaches.

\noindent \textbf{(II) Model-centric Compression ($\mathbf{W}$):} Reducing parameter complexity lowers computational and memory costs. Model compression is \textbf{\emph{model-centric}}, transforming $\mathbf{W}$ to a smaller $\mathbf{W}'$:
\begin{equation}
\mathbf{W}' = \boldsymbol{\Gamma}(\mathbf{W}), \quad \text{where} \quad |\mathbf{W}'| < |\mathbf{W}|
\end{equation}
with $\boldsymbol{\Gamma}$ as the compression operator. Key methods include: \textbf{(i) network pruning}~\cite{Network_Pruning,cheng2024survey}; \textbf{(ii) quantization}~\cite{yang2019quantization,rokh2023comprehensive}; \textbf{(iii) knowledge distillation}~\cite{Hinton:KD,gou2021KD_Survey}; \textbf{(iv) low-rank decomposition}~\cite{yu2017low-rank,idelbayev2020low-rank}. As model sizes plateau and context lengths grow, research is shifting toward data-centric compression.

\noindent \textbf{(III) Data-centric Compression ($\mathbf{X}$):} Data-centric compression is a \textbf{\emph{data-centric}} paradigm that improves efficiency by directly reducing the volume of data processed during training or inference. It encompasses two primary strategies: \textbf{(i) dataset compression}, which selects or distills informative subsets from the training corpus, and \textbf{(ii) token compression}, which directly reduces the length of input sequences during inference. Given an input sequence $\mathbf{X}$, data-centric compression yields a compressed representation $\mathbf{X}'$:
\begin{equation}
\mathbf{X}' = \boldsymbol{\Phi}(\mathbf{X}), \quad \text{where} \quad |\mathbf{X}'| < |\mathbf{X}|
\end{equation}
with $\boldsymbol{\Phi}$ as the data compression operator. This approach complements model-centric compression and has shown strong effectiveness in vision~\cite{Rao2021:DynamicViT,Bolya2023:ToMe} and language domains~\cite{kim2021length,jiang2023llmlingua}.
\section{How Data-centric Compression Drives Efficient and Effective Models}
\label{sec:methods}

In this section, we begin with the research roadmap of data-centric compression in Section~\ref{subsec:road-map}. Then, we comprehensively analyze the benefits of data-centric compression methods during both training and inference stages in Section~\ref{subsec:targets}. Finally, we summarize their advantages in Section~\ref{subsec:advantages}.

\subsection{Research Roadmap - What Makes Data-centric Compression Work?}
\label{subsec:road-map}

\begin{figure*}
    \centering
    \includegraphics[width=1.0\textwidth]{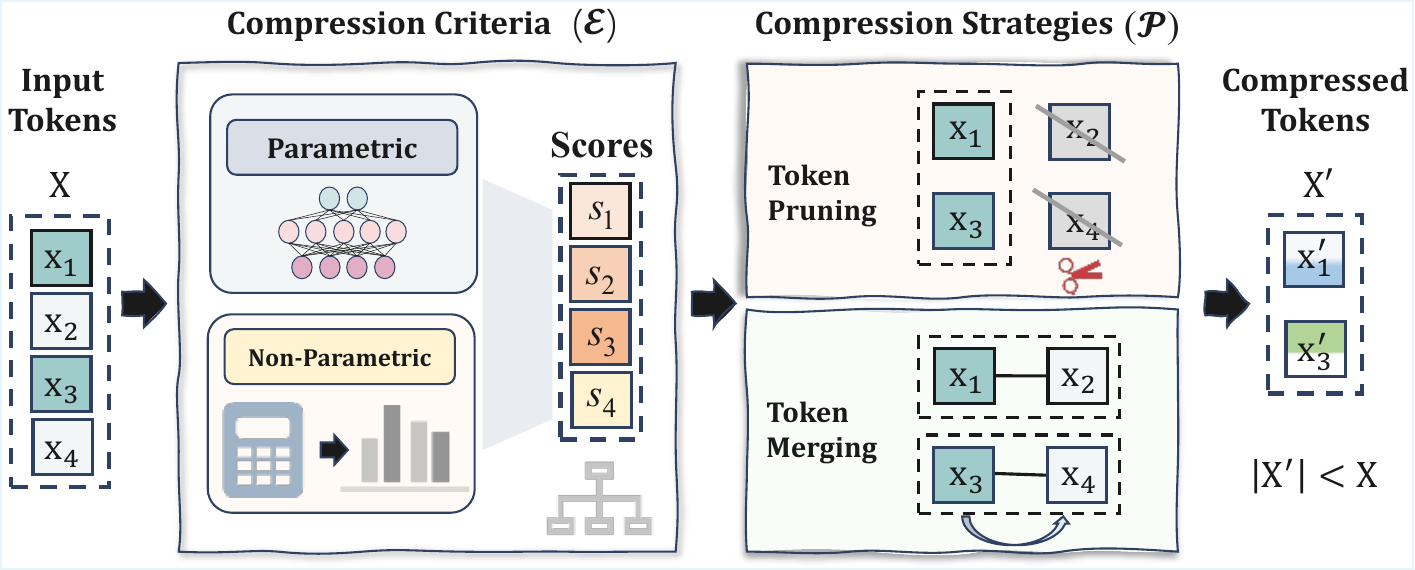}
    \vspace{-7mm}
    \caption{\textbf{Overview of the data-centric compression paradigm.} Given an input token sequence $\mathbf{X} = [\mathbf{x}_1, \dots, \mathbf{x}_T]$, data-centric compression first computes importance scores via a scoring function $\mathcal{E}: \mathbf{X} \to \{s_t\}_{t=1}^T$, then generates a compressed sequence $\mathbf{X}'$ through a compression strategy $\mathcal{P}: (\mathbf{X}, \{s_t\}_{t=1}^T) \to \mathbf{X}'$, where $|\mathbf{X}'| < |\mathbf{X}|$.}
    \label{fig:token_compression_paradigm}
    \vspace{-5mm}
\end{figure*}

Existing data-centric compression methods (\emph{i.e.}, token compression and dataset compression) fundamentally operate through a two-stage process (see Figure~\ref{fig:token_compression_paradigm}): first, identifying tokens eligible for compression within the existing token sequence $\mathbf{X} = [\mathbf{x}_1, \mathbf{x}_2, \dots, \mathbf{x}_T]$ using carefully designed \textbf{\emph{compression criteria}} through a scoring function $\mathcal{E}: \mathbf{X} \to \{s_t\}_{t=1}^T$, and then determining the precise handling of these tokens through specific \textbf{\emph{compression strategies}} $\mathcal{P}: (\mathbf{X}, \{s_t\}_{t=1}^T) \to \mathbf{X}'$ that transform the original sequence into a compressed one where $|\mathbf{X}'| < |\mathbf{X}|$. Given that existing research primarily revolves around these two key components, we next systematically analyze their designs and review representative approaches.

\paragraph{Compression Criteria ($\mathcal{E}$)} 
To determine which tokens should be compressed in sequence $\mathbf{X} = [\mathbf{x}_1, \mathbf{x}_2, \dots, \mathbf{x}_T]$, compression criteria employ scoring functions $\mathcal{E}$ to evaluate each token's importance or redundancy. Based on whether additional parameters are introduced into original models, these criteria can be categorized into two main approaches:

\noindent \textbf{(I) Parametric Methods} employ auxiliary networks as scoring functions $\mathcal{E}_{\Delta \theta}: \mathbf{X} \to \{s_t\}_{t=1}^T$, introducing additional parameters $\Delta \theta$ beyond the original model parameters $\theta$. These methods include: \textbf{(i) training-aware} approaches~\cite{Rao2021:DynamicViT,you2024DiffCR,li2024tokenpacker,kim2022learned} that optimize $\Delta \theta$ through training to learn scoring function $\mathcal{S}_{\Delta \theta}: \mathbf{X} \to \{s_t\}_{t=1}^T$, and \textbf{(ii) training-free} approaches~\cite{Mahmud2024:PaPr,zhao2024SGL} that directly employ pre-trained networks as scoring function $\mathcal{S}_{\text{fixed}}: \mathbf{X} \to \{s_t\}_{t=1}^T$ without updating $\Delta \theta$.

\noindent \textbf{(II) Non-parametric Methods} utilize parameter-free heuristics for token scoring without introducing extra parameters. These approaches can be categorized into: \textbf{(i) inherent computation} methods~\cite{Liang2022:EViT,zou2025:ToCa,Chen2024:FastV,xiao2024efficient,ge2024model} that leverage model's internal calculations for token scoring $\mathcal{S}_{\text{in}}: \mathbf{A} \to \{s_t\}_{t=1}^T$, such as using attention weights ($s_t = \sum_{j=1}^T a_t^j$, where $a_t^j$ represents attention score between tokens), and \textbf{(ii) external computation} methods~\cite{bolya2023tome,Zhang2025:SiTo,devoto-etal-2024-simple,wang2024model,liu2025vidcom2} that design additional metrics $\mathcal{S}_{\text{ex}}: \mathbf{Z} \to \mathbb{R}^{T \times T}$ to evaluate token relationships. For external methods, an additional function $g: \mathbf{X} \to \mathbf{Z}$ is introduced to compute intermediate features, where $\mathbf{Z} = g(\mathbf{X})$. The scoring function then operates on these features: $s_{i,j} = f(\mathbf{z}_i, \mathbf{z}_j)$, where $f$ is a custom pairwise scoring function. A typical example is using cosine similarity, where $g$ is an identity function and $s_{i,j} = \frac{\langle \mathbf{x}_i, \mathbf{x}_j \rangle}{\|\mathbf{x}_i\|_2 \|\mathbf{x}_j\|_2}$.

\paragraph{Compression Strategies ($\mathcal{P}$)}
To reduce sequence length while preserving critical information, compression strategies $\mathcal{P}$ transform the original sequence based on token scores $\{s_t\}_{t=1}^T$. These strategies can be categorized into two approaches:

\noindent \textbf{(I) Token Pruning} directly discards less important tokens from the sequence based on their scores. These methods~\cite{Rao2021:DynamicViT,goyal2020power,jiang2023llmlingua,Chen2024:FastV} typically remove tokens with scores below a threshold, producing a compressed sequence:
\begin{equation}
\mathbf{X}' = \mathbf{X} \setminus \{\mathbf{x}_t \mid s_t < \tau\}
\end{equation}
where $\tau$ is a threshold determining token removal. Token pruning reduces computation through direct elimination but risks information loss, particularly for fine-grained tasks~\cite{xie2021segformer}.

\noindent \textbf{(II) Token Merging} preserves information by combining semantically similar tokens~\cite{bolya2023tome,zhang2024sparsevlm,Bolya2023:ToMeSD}. Given an input sequence $\mathbf{X} = \{\mathbf{x}_1, \dots, \mathbf{x}_T\}$ and a mapping $\pi: \{1,\dots,T\} \to \{1,\dots,M\}$ that assigns tokens to $M$ merge groups based on their semantic relationships, this approach generates a compressed sequence $\mathbf{X}' = \{\mathbf{x}'_1, \dots, \mathbf{x}'_M\}$ through weighted aggregation:
\begin{equation}
\mathbf{x}'_m = \sum_{t:\pi(t)=m} w_t\mathbf{x}_t, \quad w_t = \frac{s_t}{\sum_{t':\pi(t')=m} s_{t'}}
\end{equation}
where $w_t$ represents importance weights. Token merging preserves information through weighted combinations of tokens, offering a more nuanced approach than direct elimination.

\subsection{Training and Inference Targets - How Data-centric Compression Benefits?}
\label{subsec:targets}

\paragraph{Training Stage}

Data-centric compression methods contribute to improving both the quality and efficiency of model training. These benefits can be broadly categorized into two aspects: \textit{enhancing training quality} and \textit{increasing training efficiency}.

\noindent \textbf{(I) Enhancing Training Quality}  
Improvement in training quality can be achieved through methods such as \textit{data augmentation} and \textit{token selection}, which serve to increase data diversity and emphasize the most informative content, respectively.

\noindent \textbf{(i) Data augmentation} techniques have been widely adopted to enrich training datasets by introducing variability that enhances robustness and informativeness~\cite{Autoaugment}. In computer vision, mixing or combining image tokens creates novel representations that elevate training effectiveness~\cite{mixup,cutmix}. This strategy has also been extended to synthetic datasets, where adaptive augmentation controls the informativeness of generated images~\cite{DCC,DSA,NCFM,sdc,drupi}. Analogously, in natural language processing, augmenting text tokens through synonym replacement~\cite{wei2019eda}, contraction expansion~\cite{coulombe2018text},  back-translation~\cite{chen2020mixtext}, and reformulation~\cite{hao2025maga}, supporting better generalization.

\noindent \textbf{(ii) Token selection} focuses on filtering out low-quality tokens to refine training data quality~\cite{rho1,lee2022deduplicating,penedo2023refinedweb,wenzek2019ccnet,gao2020pile,li2024datacomp,winning}. Common approaches include rule-based heuristics~\cite{raffel2020exploring,penedo2024fineweb}, deduplication methods~\cite{lee2022deduplicating,penedo2023refinedweb,abbas2023semdedup}, and scoring strategies leveraging large language models~\cite{wenzek2019ccnet,gao2020pile,li2024datacomp,wettig2024qurating,sachdeva2024train,gnothi}.

Formally, consider a training batch $\mathcal{B} = \{ \mathbf{X}_i \}_{i=1}^N$, where each $\mathbf{X}_i = [\mathbf{x}_{i,1}, \mathbf{x}_{i,2}, \dots, \mathbf{x}_{i,T}]$ is a token sequence of length $T$. A quality scoring function $q: \mathcal{T} \to \mathbb{R}$ assigns each token $\mathbf{x}_{i,j} \in \mathcal{T}$ a score reflecting its informativeness or relevance. Using a threshold $\tau$, tokens with scores below $\tau$ are filtered out via a mask $\mathbf{m}_i$:
$m_{i,j} = \begin{cases} 1, & q(\mathbf{x}_{i,j}) \geq \tau \\ 0, & \text{otherwise} \end{cases}$. The filtered batch $\tilde{\mathcal{B}}$ consists of sequences:
\begin{equation}
\tilde{\mathbf{X}}_i = \{ \mathbf{x}_{i,j} \mid m_{i,j} = 1, \quad j = 1, \ldots, T \}.
\end{equation}

Training on these curated, high-quality tokens enables the model to focus on the most relevant information, reducing noise and redundancy, thereby improving generalization and learning efficiency.

\noindent \textbf{(II) Increasing Training Efficiency}  
Data-centric compression methods directly reduce the sequence length processed during training, addressing critical challenges associated with scaling large models~\cite{Bolya2023:ToMe,Choudhury2024:RLT,Shang2024:LLaVA-PruMerge,xing2024pyramiddrop}. For Transformer architectures with sequence length reduced from $n$ to $m$ ($m < n$), the computational and memory benefits can be quantified as:
\begin{equation}
\begin{aligned}
\frac{\Omega(\mathbf{X}')}{\Omega(\mathbf{X})} &= \frac{\mathcal{O}(m^2 d)}{\mathcal{O}(n^2 d)} = \mathcal{O}\left(\frac{m^2}{n^2}\right), \\
\frac{\mathcal{M}(\mathbf{X}')}{\mathcal{M}(\mathbf{X})} &\approx \frac{md}{nd} = \frac{m}{n}.
\end{aligned}
\end{equation}
where $d$ is the embedding dimension, $\Omega(\cdot)$ represents the computational measure, and $\mathcal{M}(\cdot)$ denotes the memory measure. This quadratic reduction in computation and linear reduction in memory enables faster training iterations and larger batch sizes on fixed hardware resources.


\paragraph{Inference Stage} 
Data-centric compression methods can also enhance model inference efficiency through two key aspects: \emph{decreasing computational complexity} and \emph{reducing memory usage}.

\noindent \textbf{(I) Decreasing Computational Complexity:}
Following patterns established in training, data-centric compression achieves quadratic speedup in inference computations. Notably, many non-parametric compression methods~\cite{Bolya2023:ToMe,Chen2024:FastV} can be directly integrated into inference without additional training or architectural modifications, enabling immediate benefits across domains~\cite{zhang2024sparsevlm,wen2025stop}.

\noindent \textbf{(II) Reducing Memory Usage:}
Data-centric compression optimizes memory efficiency through two mechanisms: \textbf{(i) computing memory reduction} following the linear scaling pattern shown in training, and \textbf{(ii) KV cache optimization} for large language models~\cite{li2024snapkv,cai2024pyramidkv,wan2024look-m,wan2025meda}. During autoregressive generation, each layer caches key and value states for attention computation, with memory growing with sequence length. For a sequence of length $n$ compressed to length $m$, with $L$ layers and hidden dimension $d$, the KV cache memory reduction is:
\begin{equation}
\frac{\mathcal{M}_{\text{KV}}(\mathbf{X}')}{\mathcal{M}_{\text{KV}}(\mathbf{X})} = \frac{2Lmd}{2Lnd} = \frac{m}{n},
\end{equation}
where factor 2 accounts for both key and value states per layer.

These benefits are particularly crucial for real-time interactive systems, including UI agents~\cite{tang2025guisurvey,tang2025gui}, autonomous driving~\cite{gao2021autonomous}, and embodied AI~\cite{duan2022embodied,yang2025efficientvla}, where efficient processing of continuous inputs under resource constraints is essential.

\subsection{Compelling Advantages - Why Data-centric Compression Matters?}
\label{subsec:advantages}

Based on comprehensive analysis of data-centric compression, we identify \textbf{\emph{five compelling advantages}} that makes them particularly promising:

\begin{enumerate}
[leftmargin=10pt, topsep=0pt, itemsep=1pt, partopsep=1pt, parsep=1pt]
    \item \textbf{Universal Applicability:} Token redundancy is consistent across modalities and tasks, enabling data-centric compression in diverse settings.
    
    \item \textbf{Dual-phase Efficiency:} Data-centric compression accelerates both training and inference with minimal accuracy loss.

    \item \textbf{Architectural Compatibility:} Data-centric compression is orthogonal to compression methods and integrates seamlessly with them. It is also hardware and system friendly.

    \item \textbf{Low Implementation Costs:} Modern architectures like transformers support variable-length inputs, allowing data-centric compression without retraining or data overhead.

    \item \textbf{Quadratic Gains:} The $\mathcal{O}(n^2)$ complexity of self-attention ensures data-centric compression yields substantial computational savings.
\end{enumerate}

As AI development enters a new phase where context length becomes the primary bottleneck, the research focus of AI efficiency should shift towards data-centric compression, enabling more efficient and scalable AI systems.

\section{Current Challenges}
\label{sec:challenges}

\subsection{Performance Degradation}
\label{subsec:performance}

\begin{figure*}
    \centering
    \includegraphics[width=1.0\textwidth]{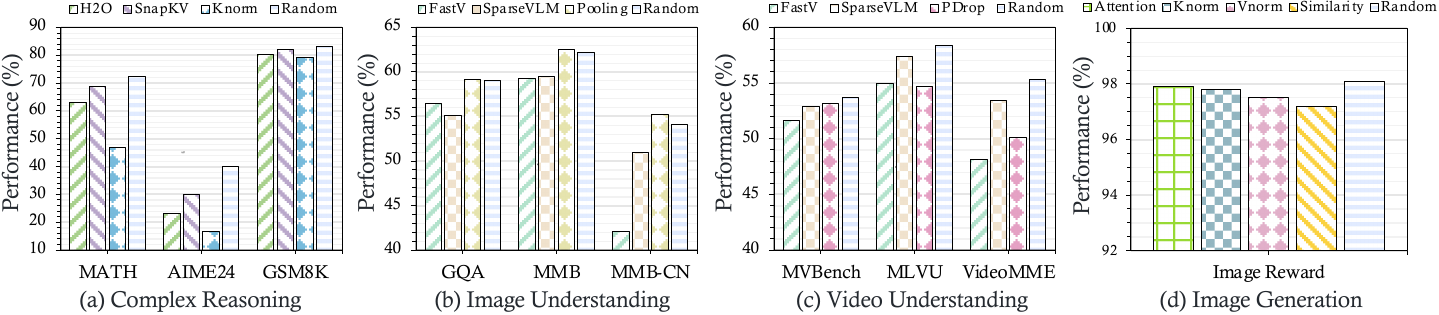}
    \vspace{-7mm}
    \caption{\textbf{Empirical comparison of carefully designed data-centric compression methods and random token dropping.} Results demonstrate that in multiple scenarios (\emph{e.g.}, LLMs, MLLMs, and DiTs), some carefully designed methods surprisingly underperform compared to random token selection.}
    \label{fig:comparison_with_random}
    \vspace{-5mm}
\end{figure*}

\noindent \textbf{Methodological Bottlenecks.}
Attention scores are central to most data-centric compression approaches. For example, \texttt{[CLS]} token attention scores are used to select key visual tokens~\cite{haurum2023tokens, zhang2024cls, han2024rethinking, yang2024visionzip, liu2025compression}, while cross-modal guidance~\cite{chen2024image, xing2024pyramiddrop, zhang2024sparsevlm} relies on text-vision attention scores.
\textbf{\emph{But are attention scores truly reliable for deciding which data to retain?}}
Recent work~\cite{zhang2024cls, wen2025token} reveals that attention scores can suffer from \textbf{position bias}.
For instance, when using text-vision scores in LLMs to retain visual tokens, those near the sequence end often receive higher weights. In 2D image space, this biases retention toward the lower half or bottom-right corner.
Clearly, it is unrealistic to assume the lower half of all images is universally more important. Such bias can significantly hurt compression performance. In Figure~\ref{fig:comparison_with_random}, this phenomenon is consistently observed across multiple tasks and models. Recent studies~\cite{jiang2025kind, wen2025stop, wen2025token, liu2025vidcom2} have also confirmed that even well-crafted attention-based methods may underperform simple random dropping. Detailed analysis is in Appendix~\ref{sec:detailed_comparison}.

\noindent \textbf{Inherent Limitations of Data-Centric Compression.}
Beyond methodological design, \textbf{\emph{does data-centric compression face inherent limitations? Is it universally applicable across tasks?}}
For MLLMs, \citeauthor{wen2025token} shows most existing methods underperform on visual grounding tasks, with significant drops on benchmarks like RefCOCO~\cite{yu2016modeling}.
In OCR-related parsing~\cite{yang2024cc, ouyang2024omnidocbench}, documents with dense layouts yield highly information-rich visual tokens. Compressing these risks severe information loss and degraded performance.
Beyond vision, current methods also face inherent limits in other modalities.
In automatic speech recognition (ASR) and automatic speech translation (AST)~\cite{ardila2019common, conneau2023fleurs}, audio is encoded and then decoded into text using an MLLM~\cite{abouelenin2025phi, chu2024qwen2}. Audio tokens are dense and temporally continuous; pruning or merging them disrupts this continuity, leading to fragmented recognition or translation.
Similarly, cross-lingual text translation may suffer significant degradation under high compression ratios.

\subsection{Suboptimal Data Representation}
\label{subsec:representation}
Most existing data-centric compression methods fall into two categories: \emph{redundancy-based} approaches that maximize information preservation between original ($\mathbf{X}$) and compressed data ($\mathbf{X'}$) via $\max_{\mathcal{C}} I(\mathbf{X}; \mathbf{X'})$, and \emph{importance-based} methods that ensure predictive sufficiency through $I(\mathbf{X'}; \mathbf{Y}) \geq I(\mathbf{X}; \mathbf{Y}) - \epsilon$, where $\epsilon$ denotes bearable information loss.
While effective for their respective objectives, we argue that these paradigms share a critical limitation: neither guarantees that the compressed data $\mathbf{X'}$ forms an \textbf{optimal representation} for downstream modeling.
The redundancy-based framework, despite preserving maximal mutual information with $\mathbf{X}$, often retains tokens with reconstructive but low discriminative value.
The importance-based framework, on the other hand, prioritizes maintaining predictive performance with respect to the target variable $\mathbf{Y}$, but often at the cost of introducing task-specific biases. By focusing solely on information relevant to a predefined label, these methods may overlook the need to maintain stable structural and semantic patterns across the sequence that could enhance generalization.
Consequently, both approaches risk producing representations misaligned with the ultimate goal of effective and generalizable downstream modeling.

\subsection{Fair Comparison}
\label{subsec:comparison}

\noindent \textbf{Rethinking FLOPs and Compression Ratios as Efficiency Metrics.}
Many data-centric compression methods report speedup by estimating FLOPs reductions or directly using token compression ratios.
\textbf{\emph{But do FLOPs or compression ratios truly reflect real acceleration?}}
Our analysis shows that, even with similar compression ratios or FLOPs, methods often vary significantly in runtime latency.
Investigating further, we find:
\textbf{(i)} Importance-based compression often uses attention scores~\cite{Chen2024:FastV, zhang2024sparsevlm}, but this can limit compatibility with efficient attention operators (\emph{e.g.}, Flash Attention~\cite{dao2023flashattention2}), contributing to the discrepancy between theoretical FLOPs and actual latency.
\textbf{(ii)} Some methods pursue high compression via progressive compression across layers~\cite{xing2024pyramiddrop}, adding computational overhead that offsets the gains from token reduction.
Thus, we argue that runtime latency should be prioritized in evaluations, as FLOP or token count reductions do not always yield real-world speedup.

\noindent \textbf{Data-centric Evaluation: The Benchmarking Gap.}
Current data-centric compression methods are mostly evaluated on general-purpose benchmarks that are not designed to capture compression-specific challenges. Consequently, benchmarks such as ScienceQA~\cite{lu2022learn} and VizWiz~\cite{bigham2010vizwiz} sometimes show improved or stable performance under compression, contradicting expectations. This suggests these benchmarks may not adequately reflect the trade-offs inherent in compression.
The issue undermines the reliability of current evaluations. Benchmarks insensitive to information loss can mask real differences between methods. Moreover, limited task diversity and the absence of compression-sensitive metrics hinder understanding of method behavior in practical settings. Without dedicated benchmarks, it is unclear whether reported gains reflect true progress or artifacts of misaligned evaluation.
\section{Future Works}
\label{sec:future_work}

\subsection{Data-Model Centric Compression Co-Development}
\label{sec:co_development}
As AI systems continue to scale in both model complexity and context length, a promising direction for future research lies in the co-development of data-centric and model-centric compression strategies. Instead of treating these approaches independently, integrating them can yield synergistic benefits—enhancing overall efficiency while maintaining, or even improving, model performance.
The most straightforward form of integration adopts a staged approach, where model-centric compression is applied first, followed by data-centric methods. For example, token compression techniques can be employed on models that have already undergone quantization, pruning, or distillation.
More advanced approaches aim for mutual reinforcement between the two paradigms. From a data-centric perspective, analyzing the layer-wise evolution of token representations may reveal that certain layers contribute minimal changes. This insight can inform model-centric compression by identifying layers suitable for removal or more aggressive quantization.
Conversely, gradient information or attention scores associated with the critical neurons retained after model pruning can also guide token selection in data-centric compression, helping to preserve only the most informative tokens.

\subsection{Dedicated Benchmarks for Data-centric Compression}
\label{sec:Dedicated_Benchmark}

Given the current limitations in evaluating data-centric compression methods using general-purpose benchmarks, we envision the development of a dedicated benchmark specifically designed to evaluate them.
Such a benchmark should comprehensively \textbf{span diverse domains}—including natural language processing, computer vision, and multi-modal tasks—and \textbf{incorporate task-specific challenges} particularly relevant to token compression, such as optical character recognition (OCR) parsing~\cite{ouyang2024omnidocbench,zhang2024ocr} and automatic speech recognition (ASR)~\cite{conneau2023fleurs,park2024long}. Furthermore, it is essential that this benchmark jointly \textbf{considers both task performance and latency}, as both are critical for real-world deployment. A well-rounded benchmark would enable a more rigorous, fair, and holistic evaluation of data-centric compression methods, ultimately driving progress in this area.

\section{Conclusion}
\label{sec:Conclusion}

In this position paper, we propose repositioning AI efficiency research by advocating a shift from model-centric to data-centric compression to address long-context processing challenges.  
We first examine recent advances in long-context capabilities across downstream scenarios, showing that performance scaling has shifted from model size to context length, underscoring the need for data-centric compression to mitigate the overhead of growing context lengths.  
We then review model efficiency approaches, with emphasis on the research roadmap of data-centric compression and its potential benefits.  
After analyzing current challenges in this area, we outline promising future directions to inspire innovation.  
Our work aims to advance AI efficiency by offering a fresh perspective and catalyzing new research.

\section{Limitations}
\label{sec:Limitations}

In this work, we review data-centric compression methods and analyze their benefits and limitations. Due to space constraints, our analysis focuses on token overhead and compression techniques in several prominent domains (\emph{e.g.}, LLMs, MLLMs, and AIGC). We acknowledge that other application areas, such as computer vision, autonomous driving, embodied intelligence, and audio/speech processing, also face growing efficiency challenges and may benefit from data-centric compression. More comprehensive cross-domain review and analysis are left for our future work.

\bibliography{reference}

\clearpage
\appendix

In the appendix, we provide additional statistical analysis of token overhead in Section~\ref{sec:llm_trends}, details of the empirical experiments in Figure~\ref{fig:comparison_with_random} in Section~\ref{sec:detailed_comparison}, and analysis of alternative positions in Section~\ref{sec:alternative_positions}.

\section{Trends in LLM Scaling: Parameters vs. Context Length}
\label{sec:llm_trends}

In this section, we provide a comprehensive analysis of the temporal progression of mainstream LLMs, documenting the growth trends in both parameter counts and context lengths. This analysis provides empirical support for our central thesis regarding the shift in computational bottlenecks from model parameters to context processing. As shown in Tables below, both in text and vision domains, model size growth has significantly slowed, while context length continues to increase. This trend indicates that the focus of research for efficient AI is shifting from model-centric compression to data-centric compression.

\section{Comparison of Token Compression Methods and Random Token Dropping}
\label{sec:detailed_comparison}

This section presents a detailed comparison between designed token compression methods and random token dropping, the simplest baseline. The analysis supports arguments in Section~\ref{subsec:performance}, showing that current token compression techniques have performance limitations. Experiments span multiple domains: complex reasoning in language, image and video understanding in vision, and text-to-image generation in AI content creation. This broad evaluation assesses token compression effectiveness across diverse tasks and modalities. Results highlight the need for more robust approaches in LLMs, MLLMs, VideoLLMs, and DiTs, emphasizing the importance of universally applicable compression strategies.

\paragraph{LLMs: Complex Reasoning}
We evaluated DeepSeek-R1-Distill-Llama-8B~\cite{guo2025deepseek-r1} on complex reasoning tasks: MATH-500~\cite{lightman2023lets}, AIME24~\cite{jia2024aime}, and GSM8K~\cite{cobbe2021training}. During decoding, we enforced a fixed KV cache budget (\emph{i.e.}, 1024 tokens) and applied KV cache evition methods: H2O~\cite{zhang2023ho}, SnapKV~\cite{li2024snapkv}, KNorm~\cite{devoto-etal-2024-simple}, and random eviction at regular intervals (\emph{i.e.}, every 512 tokens).

Figure~\ref{fig:comparison_with_random} (a) shows a counterintuitive result: H2O, SnapKV, and KNorm consistently underperform random dropping on math reasoning tasks. On AIME24, random dropping surpasses SnapKV by 10\% accuracy. These findings yield two insights: \textbf{(i)} Random dropping should be a standard baseline in KV cache studies, as it is often overlooked despite strong performance; \textbf{(ii)} Its effectiveness may stem from preserving token distribution uniformity during auto-regressive decoding, better maintaining semantic coherence than deterministic strategies. This challenges the assumption that complex policies are inherently superior and reveals gaps in current token importance modeling.

\paragraph{MLLMs: Image Understanding}
We tested LLaVA-1.5-7B~\cite{liu2023llava} on image benchmarks including GQA~\cite{hudson2019gqa} and MMB~\cite{liu2025mmbench}. We retained 25\% of visual tokens, following official LLaVA evaluation scripts\footnote{\url{https://github.com/haotian-liu/LLaVA}}. We compared FastV~\cite{Chen2024:FastV}, SparseVLM~\cite{zhang2024sparsevlm}, random dropping, and pooling.

Figure~\ref{fig:comparison_with_random} (b) shows that random dropping and pooling outperform some designed methods. We attribute this to their shared property of \textbf{spatial uniformity}, which mitigates position bias (Sec.~\ref{sec:challenges}) in attention-based methods like FastV. This also highlights the negative impact of position bias in attention scores. We thus advocate for spatial uniformity as a key design principle in token compression.

\paragraph{VideoLLMs: Video Understanding} 
We evaluated LLaVA-OneVision-7B~\cite{li2024llava-ov} on video benchmarks: MVBench~\cite{li2024mvbench}, LongVideoBench~\cite{wu2024longvideobench}, MLVU~\cite{zhou2024mlvu}, and VideoMME~\cite{fu2024videomme}. We compared FastV~\cite{Chen2024:FastV}, SparseVLM~\cite{zhang2024sparsevlm}, and PDrop~\cite{xing2024pyramiddrop} with $15\%$ visual tokens, using the LMMs-Eval framework~\cite{zhang2024lmms}\footnote{\url{https://github.com/LLaVA-VL/LLaVA-NeXT/blob/main/docs/LLaVA_OneVision.md}}.

Figure~\ref{fig:comparison_with_random} (c) shows that even with only 15\% tokens, random dropping outperforms designed methods. This implies: \textbf{(i)} Random dropping must be included as a baseline; \textbf{(ii)} VideoLLM compression should prioritize \textbf{uniform spatial and temporal token distribution} for comprehensive video representation. We hypothesize that random dropping succeeds by inherently maintaining this uniformity.

\paragraph{DiTs: Image Generation}
We tested the DiT-based model FLUX.1-dev~\cite{flux2024} with the ToCa~\cite{zou2025:ToCa} method, setting cache cycle length $N = 4$ and ratio $R = 90\%$ (10\% tokens computed per step). We compared attention-based, Key Norm (Knorm), Value Norm (Vnorm), and random selection. Surprisingly, all characteristic-based strategies yielded lower Image Reward scores than random selection.

Figure~\ref{fig:comparison_with_random} (d) confirms random selection performs well in image generation. To investigate, we designed a similarity-based strategy: select 1\% tokens randomly as base, then choose 9\% most similar to them. This led to clustered, homogeneous tokens and the worst generation quality. This suggests redundancy among similar tokens, while random selection benefits from diversity, enabling richer information representation.

\section{Alternative Positions}\label{sec:alternative_positions}

While this paper promotes data-centric compression as a key strategy for advancing Efficient AI, it is equally important to recognize and engage with alternative viewpoints that challenge the feasibility, necessity, or overall effectiveness of this approach.

\subsection{Model-Centric Compression as a Superior Alternative}
Model-centric compression methods, such as \textit{pruning}~\cite{Network_Pruning,JMLR:v12:huang11b,Deep_Compression,Rao2021:DynamicViT,jang2017Gumbel-Softmax,Pan2021:IA-RED2}, \textit{quantization}~\cite{rokh2023comprehensive,Wang:HAQ,zhou2018adaptive,yang2019quantization,Lin:AWQ,Frantar:GPTQ}, and \textit{knowledge distillation}~\cite{Hinton:KD,zhang2019self_distillation,zhang2021self_distillation,park2019relational_distillation}, have long been established as effective techniques for reducing model size and computational cost. Proponents argue that this paradigm is reliable for deployment in resource-constrained environments and maintains performance consistency. For example, pruning techniques such as DynamicViT~\cite{Rao2021:DynamicViT} dynamically remove uninformative tokens during inference, reducing the computational load by up to 30–40\% with minimal impact on accuracy. Proponents of this view claim that this approach achieves substantial speedups without discarding any original data. In contrast, data-centric methods that prune input tokens risk removing critical contextual information, which may degrade performance.

\noindent \textbf{Counterargument.}
Although model-centric compression is effective, it faces scalability issues as models and datasets grow, requiring costly full retraining and processing of entire inputs. In contrast, data-centric compression reduces input complexity upfront, easing computational burdens. Some data-centric methods update only a small parameter subset~\cite{Zhao2024:DyT,liu2024Sparse-Tuning,Lee2024:DETM}, while others enable training-free deployment~\cite{bolya2023tome,Chen2024:FastV,li2024snapkv}. Combining both approaches can improve efficiency without sacrificing accuracy~\cite{azeemi2023data,kousar2025pruning}, making data-centric methods a complement to model-centric techniques.

\subsection{Advanced Model Architectures as a more Promising Direction}
Another argument against data-centric compression is the continued advancement of model architectures that can inherently handle large datasets and long sequences more efficiently~\cite{gu2021efficiently,goldstein2024goldfinch,peng2023rwkv}. The development of transformer-based architectures, such as Vision Transformers~\cite{ViT}, Swin Transformers~\cite{liu2021swin}, and large language models like GPT-3~\cite{brown2020language}, has shown significant improvements in both accuracy and scalability. These architectures integrate advanced techniques, such as hierarchical processing, self-attention mechanisms, and dynamic sparsity, enabling them to process large amounts of data efficiently. For example, Swin Transformers~\cite{liu2021swin} utilize a window-based self-attention mechanism, which reduces the computational complexity of the standard attention mechanism, making it feasible to scale models to much larger datasets and sequences. Proponents of this view argue that as these advanced models continue to evolve, there may be less need for aggressive input compression, as these models are inherently better equipped to handle large-scale data directly.

\noindent \textbf{Counterargument.}
Advanced model architectures offer strong performance but demand substantial computational resources, especially during training~\cite{llama4,yang2025Qwen3,OpenAI:GPT-4}. Data-centric compression reduces computational load early by simplifying input data, enabling more efficient training and inference without sacrificing accuracy. Techniques like token pruning and augmentation preserve or improve performance by focusing on informative data. Combined with advanced architectures, data-centric methods enhance efficiency and maintain high performance~\cite{gao2020pile,wettig2024qurating}, making them complementary rather than competitive.

\begin{table*}[!t]
\centering
{\fontsize{9}{11}\selectfont 
\renewcommand{\arraystretch}{1.0} 
\setlength\tabcolsep{12pt} 
\begin{tabular}{lcccc}
\textbf{Model Name} & \textbf{Release Date} & \textbf{Parameters} & \textbf{Maximum Context Length} & \textbf{Model Link} \\
\shline
Qwen-1.8B            & Nov 30, 2023          & 1.8B                 & 32K                             & \href{https://huggingface.co/Qwen/Qwen-1_8B}{link} \\
Qwen-7B              & Aug 3, 2023           & 7B                   & 2K (Original), 8K (Updated)     & \href{https://huggingface.co/Qwen/Qwen-7B}{link} \\
Qwen-14B             & Sep 25, 2023          & 14B                  & 8K                              & \href{https://huggingface.co/Qwen/Qwen-14B}{link} \\
\rowcolor[rgb]{ .949,  .949,  .949}
Qwen-72B             & Nov 30, 2023          & 72B                  & 32K                             & \href{https://huggingface.co/Qwen/Qwen-72B}{link} \\
\hline
Qwen1.5-0.5B         & Early 2024            & 0.5B                 & 32K                             & \href{https://huggingface.co/Qwen/Qwen1.5-0.5B}{link} \\
Qwen1.5-1.8B         & Early 2024            & 1.8B                 & 32K                             & \href{https://huggingface.co/Qwen/Qwen1.5-1.8B}{link} \\
Qwen1.5-4B           & Early 2024            & 4B                   & 32K                             & \href{https://huggingface.co/Qwen/Qwen1.5-4B}{link} \\
Qwen1.5-7B           & Early 2024            & 7B                   & 32K                             & \href{https://huggingface.co/Qwen/Qwen1.5-7B}{link} \\
Qwen1.5-14B          & Early 2024            & 14B                  & 32K                             & \href{https://huggingface.co/Qwen/Qwen1.5-14B}{link} \\
Qwen1.5-32B          & Early 2024            & 32B                  & 32K                             & \href{https://huggingface.co/Qwen/Qwen1.5-32B}{link} \\
Qwen1.5-72B          & Early 2024            & 72B                  & 32K                             & \href{https://huggingface.co/Qwen/Qwen1.5-72B}{link} \\
Qwen1.5-110B         & Early 2024            & 110B                 & 32K                             & \href{https://huggingface.co/Qwen/Qwen1.5-110B}{link} \\
\rowcolor[rgb]{ .949,  .949,  .949}
Qwen1.5-MoE-A2.7B    & Mar 28, 2024          & 14B                  & 32K                             & \href{https://huggingface.co/Qwen/Qwen1.5-MoE-A2.7B}{link} \\
\hline
Qwen2-0.5B           & Jun 6, 2024           & 0.5B                 & 32K                             & \href{https://huggingface.co/Qwen/Qwen2-0.5B}{link} \\
Qwen2-1.5B           & Jun 6, 2024           & 1.5B                 & 32K                             & \href{https://huggingface.co/Qwen/Qwen2-1.5B}{link} \\
Qwen2-7B             & Jun 6, 2024           & 7B                   & 32K (Base), 131K (Instruct)     & \href{https://huggingface.co/Qwen/Qwen2-7B}{link} \\
Qwen2-57B-A14B       & Jun 6, 2024           & 57B                  & 32K (Base), 64K (Instruct)      & \href{https://huggingface.co/Qwen/Qwen2-57B-A14B}{link} \\
\rowcolor[rgb]{ .949,  .949,  .949}
Qwen2-72B            & Jun 6, 2024           & 72B                  & 32K (Base), 131K (Instruct)     & \href{https://huggingface.co/Qwen/Qwen2-72B}{link} \\
\hline
Qwen2.5-0.5B         & Sep 19, 2024          & 0.5B                 & 32K                             & \href{https://huggingface.co/Qwen/Qwen2.5-0.5B}{link} \\
Qwen2.5-1.5B         & Sep 19, 2024          & 1.5B                 & 32K                             & \href{https://huggingface.co/Qwen/Qwen2.5-1.5B}{link} \\
Qwen2.5-3B           & Sep 19, 2024          & 3B                   & 32K                             & \href{https://huggingface.co/Qwen/Qwen2.5-3B}{link} \\
Qwen2.5-7B           & Sep 19, 2024          & 7B                   & 128K                            & \href{https://huggingface.co/Qwen/Qwen2.5-7B}{link} \\
Qwen2.5-14B          & Sep 19, 2024          & 14B                  & 128K                            & \href{https://huggingface.co/Qwen/Qwen2.5-14B}{link} \\
Qwen2.5-32B          & Sep 19, 2024          & 32B                  & 128K                            & \href{https://huggingface.co/Qwen/Qwen2.5-32B}{link} \\
Qwen2.5-72B          & Sep 19, 2024          & 72B                  & 128K                            & \href{https://huggingface.co/Qwen/Qwen2.5-72B}{link} \\
Qwen2.5-7B-Instruct-1M & Jan 2025            & 7B                   & 1M                              & \href{https://huggingface.co/Qwen/Qwen2.5-7B-Instruct-1M}{link} \\
\rowcolor[rgb]{ .949,  .949,  .949}
Qwen2.5-14B-Instruct-1M & Jan 2025           & 14B                  & 1M                              & \href{https://huggingface.co/Qwen/Qwen2.5-14B-Instruct-1M}{link} \\
\hline
Qwen3-0.6B           & Apr 29, 2025          & 0.6B                 & 32K                             & \href{https://huggingface.co/Qwen/Qwen3-0.6B}{link} \\
Qwen3-1.7B           & Apr 29, 2025          & 1.7B                 & 32K                             & \href{https://huggingface.co/Qwen/Qwen3-1.7B}{link} \\
Qwen3-4B             & Apr 29, 2025          & 4B                   & 32K                             & \href{https://huggingface.co/Qwen/Qwen3-4B}{link} \\
Qwen3-8B             & Apr 29, 2025          & 8B                   & 131K                            & \href{https://huggingface.co/Qwen/Qwen3-8B}{link} \\
Qwen3-14B            & Apr 29, 2025          & 14B                  & 131K                            & \href{https://huggingface.co/Qwen/Qwen3-14B}{link} \\
Qwen3-32B            & Apr 29, 2025          & 32B                  & 131K                            & \href{https://huggingface.co/Qwen/Qwen3-32B}{link} \\
Qwen3-30B-A3B        & Apr 29, 2025          & 30B                  & 131K                            & \href{https://huggingface.co/Qwen/Qwen3-30B-A3B}{link} \\
\rowcolor[rgb]{ .949,  .949,  .949}
Qwen3-235B-A22B      & Apr 29, 2025          & 235B                 & 131K                            & \href{https://huggingface.co/Qwen/Qwen3-235B-A22B}{link} \\
\end{tabular}
\vspace{-2mm}
\caption{\textbf{Qwen series model specifications.} Details include release dates, parameter counts, maximum context lengths, and Hugging Face links.}
}
\label{tab:qwen_specs}
\end{table*}

\begin{table*}[t!]
\centering

\renewcommand{\arraystretch}{1.0}
\setlength\tabcolsep{12pt}
{\fontsize{8}{10}\selectfont 
\begin{tabular}{lcccc}
\textbf{Model Name} & \textbf{Release Date} & \textbf{Parameters} & \textbf{Context Length} & \textbf{Model Link} \\
\shline
DeepSeek-Coder        & November 2, 2023      & 1.3B/6.7B/33B        & 16K tokens              & \href{https://huggingface.co/deepseek-ai/deepseek-coder}{link} \\
DeepSeek-LLM          & November 29, 2023     & 7B                   & 4096 tokens             & \href{https://huggingface.co/deepseek-ai/deepseek-llm-7b-base}{link} \\
DeepSeek-LLM          & November 29, 2023     & 67B                  & 4096 tokens             & \href{https://huggingface.co/deepseek-ai/deepseek-llm-67b-base}{link} \\
DeepSeekMoE           & January 11, 2024      & 16B total, 2.7B activated & 4096 tokens         & \href{https://huggingface.co/deepseek-ai/deepseek-moe-16b-base}{link} \\
DeepSeek-Math         & April 2024            & 7B                   & 4096 tokens             & \href{https://huggingface.co/deepseek-ai/deepseek-math-7b-base}{link} \\
DeepSeek-V2           & May 6, 2024           & 236B total, 21B activated & 128K tokens        & \href{https://huggingface.co/deepseek-ai/deepseek-v2}{link} \\
DeepSeek-V2-Lite      & May 16, 2024          & 16B total, 2.4B activated & 32K tokens         & \href{https://huggingface.co/deepseek-ai/deepseek-v2-lite}{link} \\
DeepSeek-Coder-V2     & June 17, 2024         & 236B total, 21B activated & 128K tokens        & \href{https://huggingface.co/deepseek-ai/deepseek-coder-v2}{link} \\
DeepSeek-Coder-V2-Lite& June 17, 2024         & 16B total, 2.4B activated & 128K tokens        & \href{https://huggingface.co/deepseek-ai/deepseek-coder-v2-lite}{link} \\
DeepSeek-V2.5         & September 2024        & 236B total, 21B activated & 128K tokens        & \href{https://huggingface.co/deepseek-ai/deepseek-v2.5}{link} \\
DeepSeek-V3           & December 26, 2024     & 671B total, 37B activated & 128K tokens        & \href{https://huggingface.co/deepseek-ai/deepseek-v3}{link} \\
DeepSeek-R1-Zero      & January 20, 2025      & 671B total, 37B activated & 128K tokens        & \href{https://huggingface.co/deepseek-ai/deepseek-r1-zero}{link} \\
DeepSeek-R1           & January 20, 2025      & 671B total, 37B activated & 128K tokens        & \href{https://huggingface.co/deepseek-ai/deepseek-r1}{link} \\
DeepSeek-R1-Distill   & January 20, 2025      & 1.5B, 7B, 8B, 14B, 32B, 70B & 32K tokens       & \href{https://huggingface.co/deepseek-ai/deepseek-r1-distill}{link} \\
\rowcolor[rgb]{ .949,  .949,  .949}
DeepSeek-V3-0324      & March 2025            & 671B total, 37B activated & 128K tokens        & \href{https://huggingface.co/deepseek-ai/deepseek-v3-0324}{link} \\
\end{tabular}
\vspace{-2mm}
\caption{\textbf{DeepSeek series model specifications.} Details include release dates, parameter counts, and maximum context lengths.}
\label{tab:deepseek_specs}
}
\end{table*}

\begin{table*}[t!]
\centering

{\fontsize{9}{11}\selectfont
\renewcommand{\arraystretch}{1.0}
\setlength\tabcolsep{12pt}
\begin{tabular}{lcccc}
\textbf{Model Name} & \textbf{Release Date} & \textbf{Parameters} & \textbf{Context Length} & \textbf{Model Link} \\
\shline
Llama 1 7B           & February 24, 2023     & 7B                   & 2,048 tokens            & \href{https://github.com/meta-llama/llama}{link} \\
Llama 1 13B          & February 24, 2023     & 13B                  & 2,048 tokens            & \href{https://github.com/meta-llama/llama}{link} \\
Llama 1 33B          & February 24, 2023     & 33B                  & 2,048 tokens            & \href{https://github.com/meta-llama/llama}{link} \\
\rowcolor[rgb]{ .949,  .949,  .949}
Llama 1 65B          & February 24, 2023     & 65B                  & 2,048 tokens            & \href{https://github.com/meta-llama/llama}{link} \\
\hline
Llama 2 7B           & July 18, 2023         & 7B                   & 4,096 tokens            & \href{https://huggingface.co/meta-llama/Llama-2-7B}{link} \\
Llama 2 13B          & July 18, 2023         & 13B                  & 4,096 tokens            & \href{https://huggingface.co/meta-llama/Llama-2-13B}{link} \\
\rowcolor[rgb]{ .949,  .949,  .949}
Llama 2 70B          & July 18, 2023         & 70B                  & 4,096 tokens            & \href{https://huggingface.co/meta-llama/Llama-2-70B}{link} \\
\hline
Llama 3 8B           & April 18, 2024        & 8B                   & 8,192 tokens            & \href{https://huggingface.co/meta-llama/Meta-Llama-3-8B}{link} \\
Llama 3 70B          & April 18, 2024        & 70B                  & 8,192 tokens            & \href{https://huggingface.co/meta-llama/Meta-Llama-3-70B}{link} \\
\hline
Llama 3.1 8B         & July 23, 2024         & 8B                   & 128,000 tokens          & \href{https://huggingface.co/meta-llama/Meta-Llama-3.1-8B}{link} \\
Llama 3.1 70B        & July 23, 2024         & 70B                  & 128,000 tokens          & \href{https://huggingface.co/meta-llama/Meta-Llama-3.1-70B}{link} \\
\rowcolor[rgb]{ .949,  .949,  .949}
Llama 3.1 405B       & July 23, 2024         & 405B                 & 128,000 tokens          & \href{https://huggingface.co/meta-llama/Llama-3.1-405B-Instruct}{link} \\
\hline
\rowcolor[rgb]{ .949,  .949,  .949}
Llama 4 Scout        & April 5, 2025         & 109B total / 17B active & 10M tokens          & \href{https://huggingface.co/meta-llama/Llama-4-Scout-17B-16E}{link} \\
\rowcolor[rgb]{ .949,  .949,  .949}
Llama 4 Maverick     & April 5, 2025         & 400B total / 17B active & 1M tokens           & \href{https://huggingface.co/meta-llama/Llama-4-Maverick-17B-128E}{link} \\
\end{tabular}
\caption{\textbf{Llama series model specifications.} Details include release date, parameter count, context length, and Hugging Face model link.}
\vspace{-2mm}
}
\label{tab:Llama_specs}
\end{table*}

\begin{table*}[t!]
\centering
{\fontsize{9}{11}\selectfont
\renewcommand{\arraystretch}{1.0}
\setlength\tabcolsep{16pt}
\begin{tabular}{lcccc}
\textbf{Model Name} & \textbf{Release Date} & \textbf{Parameters} & \textbf{Context Length} & \textbf{Model Link} \\
\shline
GLM-130B             & August 2022           & 130B                 & 2,048 tokens            & \href{https://github.com/THUDM/GLM-130B}{link} \\
ChatGLM-6B           & March 14, 2023        & 6.2B                 & 2,048 tokens            & \href{https://huggingface.co/THUDM/chatglm-6b}{link} \\
\hline
ChatGLM2-6B          & June 25, 2023         & 6.2B                 & 32,768 tokens           & \href{https://huggingface.co/THUDM/chatglm2-6b}{link} \\
ChatGLM2-6B-32K      & July 2023             & 6.2B                 & 32,768 tokens           & \href{https://huggingface.co/THUDM/chatglm2-6b-32k}{link} \\
\hline
ChatGLM3-6B          & October 2023          & 6.2B                 & 8,192 tokens            & \href{https://huggingface.co/THUDM/chatglm3-6b}{link} \\
ChatGLM3-6B-32K      & October 2023          & 6.2B                 & 32,768 tokens           & \href{https://huggingface.co/THUDM/chatglm3-6b-32k}{link} \\
ChatGLM3-6B-128K     & November 2023         & 6.2B                 & 131,072 tokens          & \href{https://huggingface.co/THUDM/chatglm3-6b-128k}{link} \\
\hline
GLM-4-9B             & May 2024              & 9B                   & 8,192 tokens            & \href{https://huggingface.co/THUDM/glm-4-9b}{link} \\
GLM-4-9B-Chat        & May 2024              & 9B                   & 131,072 tokens          & \href{https://huggingface.co/THUDM/glm-4-9b-chat}{link} \\
\rowcolor[rgb]{ .949,  .949,  .949}
GLM-4-9B-Chat-1M     & May 2024              & 9B                   & 1,048,576 tokens        & \href{https://huggingface.co/THUDM/glm-4-9b-chat-1m}{link} \\
\end{tabular}
\caption{\textbf{GLM series model specifications.} Details include release dates, parameter counts, maximum context lengths, and Hugging Face links.}
\vspace{-2mm}
}
\label{tab:glm_specs}
\end{table*}

\begin{table*}[t!]
\centering
{\fontsize{9}{11}\selectfont
\renewcommand{\arraystretch}{1.0}
\setlength\tabcolsep{12pt}
\begin{tabular}{lcccc}
\textbf{Model Name} & \textbf{Release Date} & \textbf{Parameters} & \textbf{Context Length} & \textbf{Model Link} \\
\shline
InternLM-7B          & July 2023             & 7B                   & 8,000 tokens            & \href{https://huggingface.co/internlm/internlm-7b}{link} \\
InternLM-7B-Chat v1.1 & August 22, 2023      & 7B                   & 8,000 tokens            & \href{https://huggingface.co/internlm/internlm-chat-7b}{link} \\
\rowcolor[rgb]{ .949,  .949,  .949}
InternLM-20B         & September 20, 2023    & 20B                  & 16,000 tokens           & \href{https://huggingface.co/internlm/internlm-20b}{link} \\
\rowcolor[rgb]{.949,.949,.949}
InternLM-20B-Chat    & September 20, 2023    & 20B                  & 16,000 tokens           & \href{https://huggingface.co/internlm/internlm-chat-20b}{link} \\
\hline
InternLM2-7B         & January 17, 2024      & 7B                   & 200,000 tokens          & \href{https://huggingface.co/internlm/internlm2-7b}{link} \\
\rowcolor[rgb]{.949,.949,.949}
InternLM2-20B        & January 17, 2024      & 20B                  & 200,000 tokens          & \href{https://huggingface.co/internlm/internlm2-20b}{link} \\
\hline
InternLM2.5-7B       & July 3, 2024          & 7B                   & 200,000 tokens          & \href{https://huggingface.co/internlm/internlm2_5-7b}{link} \\
InternLM2.5-7B-Chat-1M & July 2024           & 7B                   & 1,000,000 tokens        & \href{https://huggingface.co/internlm/internlm2_5-7b-chat}{link} \\
InternLM2.5-1.8B     & August 1, 2024        & 1.8B                 & 200,000 tokens          & \href{https://huggingface.co/internlm/internlm2_5-1_8b}{link} \\
\rowcolor[rgb]{.949,.949,.949}
InternLM2.5-20B      & August 1, 2024        & 20B                  & 200,000 tokens          & \href{https://huggingface.co/internlm/internlm2_5-20b}{link} \\
\hline
\rowcolor[rgb]{.949,.949,.949}
InternLM3-8B-Instruct & January 15, 2025     & 8B                   & 32,768 tokens           & \href{https://huggingface.co/internlm/internlm3-8b-instruct}{link} \\
\end{tabular}
\caption{\textbf{InternLM series model specifications.} Details include release dates, parameter counts, maximum context lengths, and Hugging Face links.}
\vspace{-2mm}
}
\label{tab:internlm_specs}
\end{table*}

\begin{table*}[t!]
\centering

{\fontsize{9}{11}\selectfont
\renewcommand{\arraystretch}{1.0}
\setlength\tabcolsep{0.3pt}
\scalebox{0.80}{
\begin{tabular}{lcccccc}
\textbf{Model Name} & \textbf{Release Date} & \textbf{LLM Backbone} & \textbf{Max Context} & \textbf{Image Resolution} & \textbf{Max Tokens} & \textbf{Model Link} \\
\shline
LLaVA-7B            & April 2023     & Vicuna-7B          & 2K    & 224×224                   & 256        & \href{https://github.com/haotian-liu/LLaVA}{link} \\
LLaVA-13B           & April 2023     & Vicuna-13B         & 2K    & 224×224                   & 256        & \href{https://github.com/haotian-liu/LLaVA}{link} \\
\hline
LLaVA-1.5-7B        & October 2023   & Vicuna-7B-v1.5     & 4K    & 336×336                   & 576        & \href{https://huggingface.co/liuhaotian/llava-v1.5-7b}{link} \\
LLaVA-1.5-13B       & October 2023   & Vicuna-13B-v1.5    & 4K    & 336×336                   & 576        & \href{https://huggingface.co/liuhaotian/llava-v1.5-13b}{link} \\
\hline
LLaVA-NeXT-7B       & January 2024   & Mistral-7B         & 8K    & 336x\{2x2,1x\{2,3,4\}, \{2,3,4\}x1\} & 2880       & \href{https://github.com/LLaVA-VL/LLaVA-NeXT}{link} \\
LLaVA-NeXT-7B       & January 2024   & Vicuna-7B-v1.5     & 4K    & 336x\{2x2,1x\{2,3,4\}, \{2,3,4\}x1\} & 2880       & \href{https://github.com/LLaVA-VL/LLaVA-NeXT}{link} \\
LLaVA-NeXT-13B      & January 2024   & Vicuna-13B-v1.5    & 4K    & 336x\{2x2,1x\{2,3,4\}, \{2,3,4\}x1\} & 2880       & \href{https://github.com/LLaVA-VL/LLaVA-NeXT}{link} \\
LLaVA-NeXT-34B      & January 2024   & Nous-Hermes-2-Yi-34B & 4K   & 336x\{2x2,1x\{2,3,4\}, \{2,3,4\}x1\} & 2880       & \href{https://github.com/LLaVA-VL/LLaVA-NeXT}{link} \\
\hline
LLaVA-OneVision-0.5B & August 2024   & Qwen2-0.5B         & 32K   & 336×336×[6,6]              & 7290(si), 8748(mi), 6272(vid) & \href{https://huggingface.co/llava-hf/llava-onevision-qwen2-0.5b-ov-hf}{link} \\
LLaVA-OneVision-7B  & August 2024    & Qwen2-7B           & 32K   & 336×336×[6,6]              & 7290(si), 8748(mi), 6272(vid) & \href{https://huggingface.co/llava-hf/llava-onevision-qwen2-7b-ov-hf}{link} \\
\rowcolor[rgb]{ .949,  .949,  .949}
LLaVA-OneVision-72B & August 2024    & Qwen2-72B          & 32K   & 336×336×[6,6]              & 7290(si), 8748(mi), 6272(vid) & \href{https://huggingface.co/llava-hf/llava-onevision-qwen2-72b-ov-hf}{link} \\
\end{tabular}
}
\caption{\textbf{LLaVA series model specifications.} Details include release dates, backbone models, context lengths, and multimodal capabilities. si: single image; mi: multiple images; vid: video.}
\vspace{-2mm}
}
\label{tab:llava_specs}
\end{table*}

\begin{table*}[t!]
\centering

{\fontsize{9}{11}\selectfont
\renewcommand{\arraystretch}{1.0}
\setlength\tabcolsep{1pt}
\scalebox{1.0}{
\begin{tabular}{lcccccc}
\textbf{Model Name} & \textbf{Release Date} & \textbf{LLM Backbone} & \textbf{Max Context} & \textbf{Image Resolution} & \textbf{Max Tokens} & \textbf{Model Link} \\
\shline
InternVL-21B         & Dec 2023   & Vicuna-7B               & 2K    & 224×224, 336×336, 448×448      & 1,024       & \href{https://huggingface.co/OpenGVLab/InternVL-14B-224px}{link} \\
InternVL-27B         & Dec 2023   & Vicuna-13B              & 2K    & 224×224, 336×336, 448×448      & 1,024       & \href{https://huggingface.co/OpenGVLab/InternVL-14B-224px}{link} \\
\hline
InternVL1.5-26B      & Apr 2024   & InternLM2-20B           & 200K  & 2688×2688                    & 8,192       & \href{https://huggingface.co/OpenGVLab/InternViT-6B-448px-V1-5}{link} \\
\hline
InternVL2.5-1B       & Dec 2024   & Qwen2.5-0.5B-Instruct   & 32K   & 2688×2688                    & 8,192       & \href{https://huggingface.co/OpenGVLab/InternVL2_5-1B}{link} \\
InternVL2.5-2B       & Dec 2024   & Internlm2.5-1.8B-chat   & 200K  & 2688×2688                    & 8,192       & \href{https://huggingface.co/OpenGVLab/InternVL2_5-2B}{link} \\
InternVL2.5-4B       & Dec 2024   & Qwen2.5-3B-Instruct     & 32K   & 2688×2688                    & 8,192       & \href{https://huggingface.co/OpenGVLab/InternVL2_5-4B}{link} \\
InternVL2.5-8B       & Dec 2024   & Internlm2.5-7B-chat     & 200K  & 2688×2688                    & 8,192       & \href{https://huggingface.co/OpenGVLab/InternVL2_5-8B}{link} \\
InternVL2.5-26B      & Dec 2024   & Internlm2.5-20B-chat    & 200K  & 2688×2688                    & 8,192       & \href{https://huggingface.co/OpenGVLab/InternVL2_5-26B}{link} \\
InternVL2.5-38B      & Dec 2024   & Qwen2.5-32B-Instruct    & 128K  & 2688×2688                    & 8,192       & \href{https://huggingface.co/OpenGVLab/InternVL2_5-38B}{link} \\
InternVL2.5-78B      & Dec 2024   & Qwen2.5-72B-Instruct    & 128K  & 2688×2688                    & 8,192       & \href{https://huggingface.co/OpenGVLab/InternVL2_5-78B}{link} \\
\hline
InternVL3-1B         & Apr 2025   & Qwen2.5-0.5B            & 32K   & 2688×2688                    & 32K         & \href{https://huggingface.co/OpenGVLab/InternVL3-1B-Instruct}{link} \\
InternVL3-2B         & Apr 2025   & Qwen2.5-1.5B            & 32K   & 2688×2688                    & 32K         & \href{https://huggingface.co/FriendliAI/InternVL3-2B}{link} \\
InternVL3-8B         & Apr 2025   & Qwen2.5-7B              & 128K  & 2688×2688                    & 32K         & \href{https://huggingface.co/OpenGVLab/InternVL3-8B}{link} \\
InternVL3-9B         & Apr 2025   & InternLM3-8B            & 32K   & 2688×2688                    & 32K         & \href{https://huggingface.co/OpenGVLab/InternVL3-9B}{link} \\
InternVL3-14B        & Apr 2025   & Qwen2.5-14B             & 128K  & 2688×2688                    & 32K         & \href{https://huggingface.co/OpenGVLab/InternVL3-14B}{link} \\
InternVL3-38B        & Apr 2025   & Qwen2.5-32B             & 128K  & 2688×2688                    & 32K         & \href{https://huggingface.co/OpenGVLab/InternVL3-38B}{link} \\
\rowcolor[rgb]{ .949,  .949,  .949}
InternVL3-78B        & Apr 2025   & Qwen2.5-72B             & 128K  & 2688×2688                    & 32K         & \href{https://huggingface.co/OpenGVLab/InternVL3-78B}{link} \\
\end{tabular}
}
\caption{\textbf{InternVL series model specifications.} Details include release dates, backbone architectures, and multimodal capabilities.}
\vspace{-2mm}
}
\label{tab:internvl_specs}
\end{table*}

\begin{table*}[t!]
\centering

{\fontsize{9}{11}\selectfont
\renewcommand{\arraystretch}{1.0}
\setlength\tabcolsep{1pt}
\scalebox{0.98}{
\begin{tabular}{lcccccc}
\textbf{Model Name} & \textbf{Release Date} & \textbf{LLM Backbone} & \textbf{Max Context} & \textbf{Image Resolution} & \textbf{Max Tokens} & \textbf{Model Link} \\
\shline
Qwen-VL-9.6B        & Aug 2023    & Qwen-7B          & 2K    & 448×448        & 1,024       & \href{https://huggingface.co/Qwen/Qwen-VL}{link} \\
\hline
Qwen2-VL-2B         & Sep 2024    & Qwen2-1.5B       & 32K   & native resolution (max=2048×2048)      & 16,384      & \href{https://huggingface.co/Qwen/Qwen2-VL-2B}{link} \\
Qwen2-VL-7B         & Sep 2024    & Qwen2-7B         & 32K   & native resolution (max=2048×2048)      & 16,384      & \href{https://huggingface.co/Qwen/Qwen2-VL-7B-Instruct}{link} \\
Qwen2-VL-72B        & Sep 2024    & Qwen2-72B        & 32K   & native resolution (max=2048×2048)      & 16,384      & \href{https://huggingface.co/Qwen/Qwen2-VL-72B}{link} \\
\hline
Qwen2.5-VL-3B       & Feb 2025    & Qwen2.5-3B       & 32K   & native resolution (max=2048×2048)      & 24,576      & \href{https://huggingface.co/Qwen/Qwen2.5-VL-3B-Instruct}{link} \\
Qwen2.5-VL-7B       & Feb 2025    & Qwen2.5-7B       & 128K  & native resolution (max=2048×2048)      & 24,576      & \href{https://huggingface.co/Qwen/Qwen2.5-VL-7B-Instruct}{link} \\
\rowcolor[rgb]{ .949,  .949,  .949}
Qwen2.5-VL-72B      & Feb 2025    & Qwen2.5-72B      & 128K  & native resolution (max=2048×2048)      & 24,576      & \href{https://huggingface.co/Qwen/Qwen2.5-VL-72B-Instruct}{link} \\
\end{tabular}
}
\caption{\textbf{Qwen-VL series model specifications.} Includes release dates, backbone architectures, and multimodal capabilities.}
\vspace{-2mm}
}
\label{tab:qwen_vl_specs}
\end{table*}

\end{document}